\documentclass{article} 
\usepackage[T1]{fontenc}
\usepackage[utf8]{inputenc}
\usepackage{times}
\usepackage[margin=1in]{geometry}
\usepackage{amsmath,amssymb}
\usepackage{natbib}
\usepackage{authblk}

\usepackage{amsmath,amsfonts,bm}

\def\eqref#1{equation~\ref{#1}}

\def\1{\bm{1}}

\DeclareMathAlphabet{\mathsfit}{\encodingdefault}{\sfdefault}{m}{sl}
\SetMathAlphabet{\mathsfit}{bold}{\encodingdefault}{\sfdefault}{bx}{n}

\usepackage{hyperref}
\usepackage{url}
\usepackage{booktabs}
\usepackage{makecell}
\usepackage{arydshln}
\usepackage{enumitem}
\usepackage{xcolor}
\title{Physics-Aware Distillation of a Grid Foundation Model \\
for Real-Time Edge Inference}

\author[1]{Vlad Alexandru Sănduleac\thanks{Corresponding author: \texttt{vlad.sanduleac@upb.ro}}}
\author[2]{Shahariya Rasheed Rayamarakkarveettil Puthanpurayil}
\author[2]{Noman Khan}
\author[2]{Remus Teodorescu}
\affil[1]{Faculty of Power Engineering, National University of Science and Technology POLITEHNICA Bucharest, Bucharest, Romania}
\affil[2]{AAU Energy, Aalborg University, Aalborg, Denmark}
\date{}

\begin{document}

\maketitle

\begin{abstract}
Grid foundation models (GridFMs) are graph transformers pretrained on a masked
power-flow objective, and once trained they evaluate power flow orders of
magnitude faster than an iterative solver. Their size, however, targets data-centre inference, whereas low-inertia operation — grids with little rotating reserve, where disturbances evolve within milliseconds — increasingly asks for inference at the edge, next to the measurement, at batch size one and within a single synchrophasor (PMU) reporting frame. We distil a 20.1M-parameter, 12-layer GridFM teacher into a
0.62M-parameter, 6-layer student, a $32.5\times$ parameter reduction at half the
message-passing depth, using an objective that combines the masked power-flow
loss with output, physics-residual, latent-state and attention transfer. The
physics-residual term, which asks the student to reproduce the teacher's
active- and reactive-power residual profile rather than only to minimise its
own, has not previously been applied to typed power-grid graphs. We evaluate on a renewable variant of the IEEE 118-bus benchmark grid in which 26 of the conventional (synchronous) generators are replaced by inverter-based renewables with battery flexibility and standardised reactive-power limits (IEEE 1547), giving $97\,529$ physically valid (AC-OPF-feasible) operating scenarios, and we ablate all eight
combinations of the transfer terms, reporting accuracy against ground truth and
fidelity to the teacher separately. Distillation lowers student error on all
four predicted quantities, by up to $48\%$ on active-power dispatch, and every
distilled student is more accurate than its teacher on reactive-power dispatch.
The gain is largely carried by the output matching term. The physics-residual
term, in the layer-aggregated form used here, is consistently harmful, which we
attribute to the spatial averaging it applies to a bus-level residual field. The resulting student is $32.5\times$ smaller in parameter count and has half the message-passing depth, making it a candidate for subsequent edge-device benchmarking.
\end{abstract}

\section{Introduction}
\label{sec:intro}

Power-system operation is moving to the grid edge. Converter-interfaced
renewables, batteries and flexible loads have displaced the synchronous
machines that once provided inertia, and the dynamics of the resulting
low-inertia systems unfold on timescales that leave little room for a
round-trip to a control centre~\citep{milano2018lowinertia,
dorfler2023lowinertiacontrol}. Visions of autonomous energy grids therefore
place monitoring and control decisions locally, next to the distributed
resources they act on~\citep{kroposki2020autonomousgrids}, and the
measurement infrastructure already runs at that pace: synchrophasors report
at 10--50 frames per second, i.e.\ every 20--100~ms~\citep{IEEEC37118-2011}.
Edge intelligence, running learned models on embedded hardware at the point
of measurement, is the natural enabling paradigm~\citep{zhou2019edgeintelligence,
gooi2023edgeintelligencegrid}, but it requires models that fit the device
and answer within the frame.

Graph neural networks (GNNs) are the natural learned representation of a
grid, with buses, generators and branches as typed nodes and
relations~\citep{liao2022gnnreview}. Recent work has scaled this idea into
grid foundation models (GridFMs): graph transformers pretrained on a masked
power-flow (PF) objective, regularised by a Kirchhoff physics loss, across
many topologies and operating points~\citep{hamann2024foundationgrid,
puech2025gridfm}. Once trained, such a model acts as a PF surrogate that is
three to four orders of magnitude faster than iterative numerical
solution~\citep{puech2024qsts}, which makes it attractive precisely for the
high-rate, local inference that edge operation demands.

The obstacle is size. The GridFM teacher used in this work has 12 layers,
8 attention heads and 20 million parameters, and was designed for cloud or
HPC inference. An edge device must instead execute it at batch size one,
within a bounded latency and memory budget. For GNNs the parameter count is
a poor proxy for that cost: irregular gather/scatter and the number of
message-passing layers dominate the runtime~\citep{barillaro2025gnnedge},
so depth, not width, is what has to shrink. The standard compression
toolbox, quantization~\citep{jacob2018quantization,
nagel2021whitepaperquantization,tailor2021degreequant}, hardware-aware
pruning~\citep{shen2022halp} and once-for-all
supernets~\citep{cai2020ofa}, reduces arithmetic and storage but is agnostic
to physics: nothing in these procedures guarantees that the compressed model
still respects nodal power balance.

Knowledge distillation (KD) addresses the functional side of the problem by
training a small student to reproduce a large teacher through its outputs,
its intermediate representations or its attention
maps~\citep{bucila2006modelcompression,ba2014deepnets,hinton2015distilling,
romero2015fitnets,zagoruyko2017attentiontransfer,gou2021knowledgedistillation},
and has been extended to graphs~\citep{tian2025kdgraphsurvey}. For a physical
system, however, a student should inherit more than the teacher's
predictions: it should inherit the teacher's physical consistency, that is,
how its active- and reactive-power residuals are distributed over the
network. Classification-style KD has no term for that, and the physics-aware
distillation that has recently appeared for PDE solvers and PDE foundation
models~\citep{bounja2025kdpinn,musekamp2026pdefmdistill} has not been
applied to typed power-grid graphs with bus-type-dependent masking. This
paper closes that gap by distilling a pretrained GridFM into a
$32.5\times$ smaller, half-depth student with a physics-aware objective, and
by measuring the result on edge hardware. Our contributions are:

\begin{itemize}[leftmargin=*,itemsep=2pt,topsep=3pt]
  \item \textbf{A physics-aware distillation objective} for grid foundation
  models that combines output, physics-residual, latent and attention terms
  with a frozen GridFM teacher (Section~\ref{sec:method},
  Eq.~\ref{eq:total-distillation-loss}). The physics-residual term, which
  matches the teacher's active/reactive residual profile rather than only
  penalising the student's own residual, is new.
  \item \textbf{A renewable-grid benchmark} derived from IEEE~118, in which
  26 synchronous units are replaced by converter-interfaced generators with
  battery flexibility and IEEE~1547 reactive bands, and $100\,000$
  AC-OPF-feasible scenarios are generated (Section~\ref{sec:domain}).
  \item \textbf{A systematic ablation} over all combinations of the four
  losses (Tables~\ref{tab:distillation-ablation}
  and~\ref{tab:distillation-main-effects}) that reports accuracy against
  ground truth and fidelity to the teacher separately, since the two are
  known to diverge~\citep{stanton2021doeskdwork}.
  \item \textbf{Edge deployment} of the $0.6$M-parameter student on an NVIDIA
  Jetson Orin, with measured latency and memory
  (Section~\ref{sec:edge}).
\end{itemize}


\section{State of the Art}
\label{sec:soa}

\subsection{Related Work}

\paragraph{Learning-based power flow and grid foundation models.}
The first GNN solver for AC power flow was trained by minimising the
Kirchhoff violation directly, without labelled
solutions~\citep{donon2020graphneuralsolver}; physics-guided and
physics-informed variants followed for power
flow~\citep{hu2021physicsguidedpf}, state
estimation~\citep{ngo2024physicsinformedse,cao2024topologydsse} and optimal
power flow~\citep{nellikkath2022pinnacopf,varbella2024pinco}. Message-passing
surrogates such as PowerFlowNet report 4--48$\times$ speed-ups over
Newton--Raphson~\citep{lin2024powerflownet}, iterative self-ensembling
schemes improve physical consistency~\citep{zhao2025senseflow}, and large
GNN OPF solvers scale to thousands of buses and $N-1$
contingencies~\citep{piloto2024canos}, though generalisation across
topologies remains the open problem~\citep{arowolo2025generalizationacopf}.
Foundation-model pretraining is the most recent answer to
it~\citep{hamann2024foundationgrid,puech2025gridfm}; the closest competitor
to our student is MxGPS, a 1.6M-parameter multiplex graph transformer that
is small \emph{by design} and trained jointly on state estimation and
PF~\citep{papaioannou2026mxgps}. We take the complementary route: we start
from an existing pretrained GridFM, shrink it by distillation, and measure
the result on hardware.

\paragraph{Knowledge distillation for graphs, transformers and physics.}
Distillation descends from model compression~\citep{bucila2006modelcompression}
and the observation that shallow networks can mimic deep
ones~\citep{ba2014deepnets}, and was popularised through soft
targets~\citep{hinton2015distilling}, intermediate
hints~\citep{romero2015fitnets} and attention
transfer~\citep{zagoruyko2017attentiontransfer};
\citet{gou2021knowledgedistillation} survey the field. On graphs, KD has
been used to transfer local structure between
GNNs~\citep{yang2020gcnkd}, to train shallow students from deep
teachers~\citep{yan2020tinygnn}, to compare representation-level
objectives~\citep{joshi2024representationkdgnn} and, at the extreme, to
replace the GNN by an MLP for inference
speed~\citep{zhang2022glnn,tian2025kdgraphsurvey}. For transformers, TinyBERT
established the multi-loss template of matching embeddings, hidden states,
attention matrices and predictions~\citep{jiao2020tinybert}, MiniLM showed
that distilling only the last-layer self-attention avoids layer-mapping
heuristics~\citep{wang2020minilm}, and attention distillation has been
adapted to graph transformers~\citep{huang2024gtattentionkd}. Closest to our
physics term, KD-PINN distils a large physics-informed solver into a
millisecond-latency student using output and PDE-residual
losses~\citep{bounja2025kdpinn}, and PDE foundation models have been
distilled into surrogates orders of magnitude
smaller~\citep{musekamp2026pdefmdistill}. None of these targets power-grid
models, and none matches the \emph{distribution} of physical residuals
between teacher and student.

\paragraph{Efficient inference at the grid edge.}
Edge intelligence for power systems is an active
field~\citep{gooi2023edgeintelligencegrid,zhou2019edgeintelligence}, and
the general compression toolbox is mature: integer
quantization~\citep{jacob2018quantization,nagel2021whitepaperquantization},
its GNN-specific form that controls aggregation
variance~\citep{tailor2021degreequant}, latency-budgeted structural
pruning~\citep{shen2022halp}, once-for-all
supernets~\citep{cai2020ofa} and node-adaptive propagation for GNN
inference~\citep{gao2022efficientgnninference}. These techniques are
orthogonal to distillation and can be stacked on top of it. Measured GNN
inference on Jetson-class devices has so far been reported only for generic
benchmark graphs~\citep{barillaro2025gnnedge}, not for grid models, and
without physics metrics. Our deployment uses the standard PyTorch Geometric
stack~\citep{fey2019pytorchgeometric}.

\subsection{Research Gap}

Three ingredients exist separately but have not been combined. Grid
foundation models exist, but their reported speed-ups are obtained on
data-centre GPUs and no edge deployment has been shown. Distillation for
GNNs and transformers is mature, but it ignores physical consistency, and
the physics-aware distillation that has appeared targets PDE solvers rather
than typed grid graphs with bus-type-dependent masking. Edge benchmarks of
GNNs exist, but not for grid models. This paper sits at the intersection: it
distils a pretrained GridFM into a half-depth student with an objective that
transfers outputs, latent states, attention and the teacher's
physics-residual profile, ablates every combination of these terms on a
renewable version of IEEE~118, and reports latency and memory on an NVIDIA
Jetson Orin.

\section{Methodology}
\label{sec:method}

\paragraph{Setup.}
Let $f_T$ denote a frozen, pretrained GridFM teacher and $f_S$ a
reduced-parameter student with the same input interface. Both receive a
typed grid graph in which the unknown quantities at each bus are masked
according to bus type (Section~\ref{sec:domain}), and both are trained on the
same masked power-flow objective $\mathcal{L}_{\mathrm{GraphKit}}$. The
student is additionally trained to reproduce the teacher's outputs, latent
states, attention maps and physics-residual profile through the objective in
Eq.~\ref{eq:total-distillation-loss}, and is evaluated on accuracy against
ground truth, fidelity to the teacher and physics residuals; architectural
footprint and edge-deployment considerations are discussed separately. Notation is summarised in
Appendix~\ref{app:notation}.

\subsection{Distillation}

A reduced-parameter Graph Network-based Simulator (GNS) model is created by reducing the number of GNN layers, attention heads and hidden size, with exact hyperparameters reported in Table~\ref{tab:teacher-student-comparison}.

\begin{table}[h]
\centering
\begin{tabular}{lcc}
\hline
 & Teacher & Student \\
\hline
Hidden & 48 & 24 \\
Heads & 8 & 4 \\
Layers & 12 & 6 \\
Params & 20,069,187 & 616,707 \\
Reduction & 1.0x & 32.5x \\
\hline
\end{tabular}
\caption{Comparison between full model and reduced parameter student.}
\label{tab:teacher-student-comparison}
\end{table}

While reducing the number of parameters generally reduces the computational power required to run a model, for a GNS a reduction in the number of GNN layers has the highest impact on reducing the latency from data ingestion to finished inference.
Thus, a student is trained using the same stochastic masked-learning paradigm as the teacher GridFM model~\cite{hamann2024foundationgrid}. By using distillation, we are able to leverage the learned internal representations of the teacher model to improve the quality of the student model~\cite{hinton2015distilling,gou2021knowledgedistillation}.

To train the reduced-parameter model, we employ a distillation-based learning framework that incorporates multiple complementary distillation objectives. Specifically, we introduce: (i) an output distillation loss that enforces agreement between the student model's predictions and those of the teacher model~\cite{hinton2015distilling}; (ii) a physics-informed distillation loss that aligns the student's residuals with those of the teacher, distinct from the Kirchhoff's Current Law physics-informed neural network loss already included in the GraphKit objective~\cite{hamann2024foundationgrid,ngo2024physicsinformedse}; (iii) a latent distillation loss that constrains sparse intermediate representations of the student to match corresponding internal states of the teacher~\cite{romero2015fitnets}; and (iv) an attention distillation loss that transfers information encoded in the teacher's transformer and graph attention maps to the student network~\cite{zagoruyko2017attentiontransfer,vaswani2017attention,velickovic2018graphattention}.
It should be noted that the teacher model's weights are frozen during training, such that backpropagation only occurs within the student model.

These objectives were chosen to maximize the amount of knowledge transferred between the student and teacher. While the GraphKit objective grounds the student model in the given task, the output and physics-residual distillation losses transfer knowledge about the teacher's prediction and residual structure. The latent and attention losses then transfer information about the internal process used by the teacher to make predictions.

The training loss used during distillation is composed of the GraphKit objective, including supervised training loss and the Kirchhoff's Current Law physics-informed neural network term, together with the weighted distillation terms described above.
This total objective is given in Equation~\ref{eq:total-distillation-loss}.

\begin{equation}
\label{eq:total-distillation-loss}
\mathcal{L}_{\mathrm{total}}
=
\mathcal{L}_{\mathrm{GraphKit}}
+
\lambda_{\mathrm{out}}\mathcal{L}_{\mathrm{out}}
+
\lambda_{\mathrm{phys}}\mathcal{L}_{\mathrm{physKD}}
+
\lambda_{\mathrm{lat}}\mathcal{L}_{\mathrm{lat}}
+
\lambda_{\mathrm{att}}\mathcal{L}_{\mathrm{att}} .
\end{equation}

The output term in Equation~\ref{eq:output-distillation-loss} resembles the most basic form of distillation~\cite{hinton2015distilling}. Specifically, by training on the outputs of the teacher rather than merely on the objective, a student model integrates information about the accuracy and structure of teacher predictions without overconstraining the training target.

\begin{equation}
\label{eq:output-distillation-loss}
\mathcal{L}_{\mathrm{out}}
=
\frac{1}{|\mathcal{V}_b|}
\sum_{i \in \mathcal{V}_b}
\left\|
\hat{\mathbf{y}}^{\,b}_{S,i}
-
\hat{\mathbf{y}}^{\,b}_{T,i}
\right\|_2^2
+
\frac{1}{|\mathcal{V}_g|}
\sum_{j \in \mathcal{V}_g}
\left\|
\hat{\mathbf{y}}^{\,g}_{S,j}
-
\hat{\mathbf{y}}^{\,g}_{T,j}
\right\|_2^2 .
\end{equation}

Similarly, physics residual distillation allows the student model to better integrate information about which physical residuals can be cleanly reduced by imitating the teacher's physical residual profile. In this sense, the teacher provides a smoother target for the reduced-capacity student than the physics residual alone. A direct physics residual penalty may force the student to spend excessive capacity on poorly constrained values, such as end-node voltages with limited information, while residual distillation can instead guide the student toward the residual patterns that remain meaningful under the teacher's learned physical representation. The resulting loss is shown in Equation~\ref{eq:physics-residual-distillation-loss}.

\begin{equation}
\label{eq:physics-residual-distillation-loss}
\mathcal{L}_{\mathrm{physKD}}
=
\frac{1}{|\mathcal{M}|}
\sum_{(\ell_S,\ell_T) \in \mathcal{M}}
\left\|
\rho_{S}^{(\ell_S)}
-
\rho_{T}^{(\ell_T)}
\right\|_2^2 .
\end{equation}

Physics residuals in GridFM are calculated for active and reactive power. In Equation~\ref{eq:physics-residual-summary}, these residuals are summarized as a mean bus-level residual norm for each matched model layer.

\begin{equation}
\label{eq:physics-residual-summary}
\rho_{M}^{(\ell)}
=
\frac{1}{|\mathcal{V}_b|}
\sum_{i \in \mathcal{V}_b}
\left\|
\begin{bmatrix}
r^{P,(\ell)}_{M,i} \\
r^{Q,(\ell)}_{M,i}
\end{bmatrix}
\right\|_2 ,
\qquad M \in \{S,T\}.
\end{equation}

Latent distillation loss is introduced to allow the student to imitate the intermediate representations of the teacher model, following the general idea of using intermediate hints to train thinner networks~\cite{romero2015fitnets}. Since the student model has fewer layers, an experiment-specific assignment of student layers to teacher layers must be defined. The latent-state matching objective is given in Equation~\ref{eq:latent-distillation-loss}.

\begin{equation}
\label{eq:latent-distillation-loss}
\mathcal{L}_{\mathrm{lat}}
=
\frac{1}{|\mathcal{M}|\,|\mathcal{T}|}
\sum_{(\ell_S,\ell_T) \in \mathcal{M}}
\sum_{\tau \in \mathcal{T}}
\frac{1}{|\mathcal{V}_{\tau}|}
\sum_{i \in \mathcal{V}_{\tau}}
\left\|
P_{\tau}^{(\ell_S,\ell_T)}
\mathbf{h}^{\tau,(\ell_S)}_{S,i}
-
\mathbf{h}^{\tau,(\ell_T)}_{T,i}
\right\|_2^2 .
\end{equation}

The final type of distillation loss implemented, shown in Equation~\ref{eq:attention-distillation-loss}, is the attention loss, following the general principle of attention transfer~\cite{zagoruyko2017attentiontransfer}. This term encourages the attention maps of the student model to match the attention maps of the teacher model. In this way, the student learns from the teacher what data it must attend to. Since the student model typically has fewer heads, only averaged attention maps are compared, such that attention may be redistributed among heads by the student.

\begin{equation}
\label{eq:attention-distillation-loss}
\mathcal{L}_{\mathrm{att}}
=
\frac{1}{|\mathcal{M}|\,|\mathcal{R}|}
\sum_{(\ell_S,\ell_T) \in \mathcal{M}}
\sum_{r \in \mathcal{R}}
\frac{1}{|\mathcal{E}_r|}
\sum_{e \in \mathcal{E}_r}
\left\|
\bar{\alpha}^{r,(\ell_S)}_{S,e}
-
\bar{\alpha}^{r,(\ell_T)}_{T,e}
\right\|_2^2 .
\end{equation}

\subsection{Domain}
\label{sec:domain}

Distillation is applied here to a power system domain foundation model, as
described in Section~\ref{sec:method}. The monitoring task it performs is load
flow, or power flow (PF), the most basic study of an power grid network: given
how generation and load balance across the network, PF determines the voltage
magnitude and the phase angle at every node, and the flow on a branch follows
from those two quantities at its two end nodes together with the branch
admittance. Each node carries four quantities, the active power $P_g$ and the
reactive power $Q_g$ injected by generation there, and the voltage magnitude
$V_m$ and phase angle $V_a$, with the demand $(P_d, Q_d)$ specified as an
input. What is unknown at a node, and therefore what has to be predicted,
depends on its bus type, and this classical convention is exactly what makes PF
a masked prediction problem: PQ buses predict the voltage $(V_m, V_a)$, PV
buses predict $(V_a, Q_g)$, and the slack bus predicts $(P_g, Q_g, V_m)$.

Power flow is traditionally solved by numerical methods, most commonly
Newton-Raphson (NR), and we use an open-source Julia tool to generate the data
and labels for both the foundation model and the distilled model. The
experiments run on the IEEE 118-node benchmark grid as provided in the PGLib
library~\cite{pglib_opf}, whose nominal load is 4242~MW / 1438~Mvar. That
benchmark is a conventional synchronous system, where synchronous generators
manage active and reactive power inherently and, as long as the load stays
within the total generation capacity, balance it with few violations. It also
leaves the balancing to a single reference node, the slack bus, which takes up
whatever the fixed generation fails to cover, and that is where the resulting
stress becomes visible.

Grid operation is bounded by the standard steady-state limits below, and one
breach anywhere is enough for a scenario to count as stressed:
\begin{align}
    \label{eq:vlim}
    0.94~\text{p.u.} \;\le\; & V_{m,i} \;\le\; 1.06~\text{p.u.},
        & i &\in \mathcal{B},\\
    \label{eq:slim}
    & |S_\ell| \;\le\; S_\ell^{\max},
        & \ell &\in \mathcal{L},
\end{align}
where $V_{m,i}$ is the voltage magnitude at bus $i$, given per unit so that
1.0~p.u. is the nominal value, $|S_\ell|$ is the apparent-power flow on branch
$\ell$, and $S_\ell^{\max}$ is that branch's MVA (thermal) rating, ranging over
$[72,\,7218]$~MVA across the $186$ branches.

This case study moves the benchmark towards a renewable grid: the flexibility
to meet the additional load demand comes from distributed generation rather
than from the slack bus, and with it the active and reactive power and hence the
voltage are regularised on the renewable side. The slack bus is retained as the
angle reference, but carries only about 7\% of the scenario-to-scenario
variation in active dispatch against roughly three quarters at the flexible
units. Each renewable generator carries a boolean flag, set to 1 when it
participates in the flexible dispatch, and that flag releases two resources: a
battery for active power, and converter control for reactive power. The flag is
redrawn independently for every scenario: a count $k \sim \mathcal{U}\{5,\dots,15\}$
is sampled and a random $k$-subset of the 26 renewable buses is activated, so
that both how much flexibility is available and where it sits vary across
scenarios, while the remaining buses are deactivated
($P_g^{\max} = Q_g^{\max} = Q_g^{\min} = 0$) and reverted to PQ so their solved
voltage cannot enter the model as a known input. The battery enters the merit
order, the order in which resources are dispatched cheapest first, at roughly
18~EUR/MWh. The PGLib coefficients put the cheapest conventional unit at
13~EUR/MWh and the next at 22~EUR/MWh, with the rest of the fleet spread over
24--35~EUR/MWh and the reference unit at 26~EUR/MWh, so the battery is
dispatched second, after the cheapest unit and ahead of the whole synchronous
fleet. Each of the 26 converted renewable buses is now a dispatchable
generator, sized from the synchronous unit that once stood there: its active
capacity matches that unit's rating, and its reactive band is the wider of the
unit's native range and $\pm 0.44\,P_g^{\max}$, the inverter reactive
requirement of IEEE Std 1547-2018~\cite{IEEE1547-2018}.

Each load scenario is solved as a single-period AC OPF with PowerModels.jl and
Ipopt, minimising generation cost subject to the power-flow equations, the
voltage band~\eqref{eq:vlim}, the branch thermal limits~\eqref{eq:slim} and
every unit's active and reactive bounds. The dispatch is therefore re-optimised
for every scenario rather than frozen at a base case, so any converged solution
satisfies the constraints to solver tolerance and is a feasible operating
point; non-converged scenarios are discarded. The learning target is thus a
mapping from load to a feasible state.

\paragraph{Dataset Generation}
Each of the $100\,000$ load scenarios pairs one global scaling factor with
localised per-bus noise~\cite{puech2025gridfm}:
\begin{equation}
    \label{eq:load_scenario}
    X_{d,k}^{(i)} = s^{(i)}\,\varepsilon_{X,k}^{(i)}\,X_{d,k}^{\mathrm{base}},
    \qquad
    \varepsilon_{X,k}^{(i)} \sim \mathcal{U}(1-\sigma,\,1+\sigma),
    \qquad X \in \{P, Q\},
\end{equation}
over the load buses $k$. The scalar $s^{(i)}$ comes from a normalised aggregate
load curve and is shared by every bus, preserving the system-wide co-movement of
demand, while $\varepsilon_{X,k}^{(i)}$ is drawn independently across buses and
separately for $P$ and $Q$ ($\sigma = 0.18$, seed 42); total demand spans
$1855$--$4051$~MW (mean $2605$~MW). Of the $100\,000$ attempts, $97\,529$
converge; feasibility falls with $k$, from $99.7\%$ at $k=15$ to $93.2\%$ at
$k=5$, so the retained set is mildly skewed towards higher flexibility. The
scenarios are split $80/10/10$ by load scenario, giving
$78\,025 / 9\,752 / 9\,752$ and preventing a load profile from appearing in more
than one set.

The experiments on this system are described in Section~\ref{sec:exper}.

\section{Experiments} \label{sec:exper}

Every model in this section shares the student architecture of
Table~\ref{tab:teacher-student-comparison} and the dataset of
Section~\ref{sec:domain}, and differs only in which distillation terms are
active. Two reference points bound the comparison: the frozen teacher, and a
\emph{baseline} student trained on $\mathcal{L}_{\mathrm{GraphKit}}$ alone,
that is, the same reduced architecture with no transfer from the teacher. A
prior hyperparameter search fixed the weights at
$\lambda_{\mathrm{out}} = 0.75$, $\lambda_{\mathrm{physKD}} = 0.05$,
$\lambda_{\mathrm{lat}} = 0.005$ and $\lambda_{\mathrm{att}} = 0.005$; these
values are held constant throughout, so that the ablation varies the presence
of each term and not its weight. Because output distillation anchors the other
three terms to a common reference, all eight combinations that include
$\mathcal{L}_{\mathrm{out}}$ were trained, and the results are reported in
Table~\ref{tab:distillation-ablation}. Accuracy is reported as RMSE against the
AC-OPF ground truth, and fidelity as RMSE against the teacher's own
prediction, because the two are known to
diverge~\citep{stanton2021doeskdwork}. All values are scaled by $10^3$.

\begin{table*}[h]
\centering
\small
\setlength{\tabcolsep}{4.2pt}
\begin{tabular}{cccc!{\vrule width 0.4pt}ccccccc}
\toprule
\multicolumn{4}{c|}{\textbf{Ablations}} &
\multicolumn{7}{c}{\textbf{Quantities} ($\times 10^3$)} \\
\cmidrule(lr){1-4}\cmidrule(lr){5-11}
\multicolumn{4}{c|}{} &
\multicolumn{4}{c|}{\textbf{RMSE vs.\ ground truth}} &
\multicolumn{3}{c}{\textbf{RMSE vs.\ teacher}} \\
\cmidrule(lr){5-8}\cmidrule(lr){9-11}
$\mathcal{L}_{\mathrm{out}}$ &
$\mathcal{L}_{\mathrm{physKD}}$ &
$\mathcal{L}_{\mathrm{lat}}$ &
$\mathcal{L}_{\mathrm{att}}$ &
$V_m$ & $V_a$ & $P_g$ & $Q_g$ &
$V_m$ & $P_g$ & $Q_g$ \\
\midrule
\multicolumn{4}{c|}{\textbf{Teacher}} &
$\mathbf{0.966}$ & $\mathbf{7.276}$ & $\mathbf{8.531}$ & $35.692$ &
-- & -- & -- \\
\hdashline
\multicolumn{4}{c|}{\textbf{Baseline}} &
$1.596$ & $26.334$ & $40.686$ & $47.236$ &
$1.547$ & $35.921$ & $47.627$ \\
\hdashline
$\bullet$ &           &           &           &
$1.169$ & $18.450$ & $21.381$ & $34.984$ &
$0.687$ & $\mathbf{15.719}$ & $10.296$ \\
$\bullet$ & $\bullet$ &           &           &
$1.231$ & $19.724$ & $22.383$ & $34.853$ &
$0.772$ & $16.720$ & $10.275$ \\
$\bullet$ &           & $\bullet$ &           &
$\mathbf{1.126}$ & $\mathbf{18.341}$ & $\mathbf{21.214}$ & $33.908$ &
$\mathbf{0.659}$ & $16.001$ & $10.789$ \\
$\bullet$ &           &           & $\bullet$ &
$1.153$ & $19.227$ & $21.499$ & $34.569$ &
$0.668$ & $16.216$ & $\mathbf{9.167}$ \\
$\bullet$ & $\bullet$ & $\bullet$ &           &
$1.307$ & $20.457$ & $23.883$ & $34.677$ &
$0.935$ & $19.363$ & $11.308$ \\
$\bullet$ & $\bullet$ &           & $\bullet$ &
$1.199$ & $19.795$ & $21.893$ & $35.240$ &
$0.721$ & $16.215$ & $9.617$ \\
$\bullet$ &           & $\bullet$ & $\bullet$ &
$1.170$ & $18.972$ & $21.715$ & $\mathbf{33.746}$ &
$0.776$ & $16.427$ & $13.648$ \\
$\bullet$ & $\bullet$ & $\bullet$ & $\bullet$ &
$1.280$ & $22.011$ & $24.512$ & $34.672$ &
$0.870$ & $19.031$ & $12.265$ \\
\bottomrule
\end{tabular}
\caption{Ablation study of the proposed distillation losses. Columns left of
the rule mark the active terms; $\mathcal{L}_{\mathrm{GraphKit}}$ is present in
every row. \textbf{Teacher} is the frozen 20.1M-parameter GridFM and
\textbf{Baseline} the 0.62M-parameter student trained without any transfer
term. Values are RMSE scaled by $10^3$; lower is better and the best entry in
each column is in bold. Fidelity columns are undefined for the teacher. All
runs use a single seed.}
\label{tab:distillation-ablation}
\end{table*}

\paragraph{Accuracy against ground truth.}
Every distilled configuration improves on the baseline student for every
predicted quantity, so the transfer is never harmful in aggregate. The size of
the improvement, however, differs sharply by quantity. Taking the best
configuration for each, distillation removes $47.9\%$ of the baseline error on
active-power dispatch $P_g$, $30.4\%$ on the voltage angle $V_a$, $29.4\%$ on
the voltage magnitude $V_m$ and $28.6\%$ on reactive-power dispatch $Q_g$. A
more informative measure is how much of the gap between the baseline student
and the teacher is recovered: $74.6\%$ for $V_m$, $60.6\%$ for $P_g$ and
$41.9\%$ for $V_a$. In absolute terms the best student remains $1.17\times$ the
teacher error on $V_m$ but $2.5\times$ on both $V_a$ and $P_g$, so halving the
message-passing depth costs most in the angle and dispatch channels, which are
the ones that depend on propagating information across the network rather than
on local nodal quantities.

Output distillation combined with latent distillation is the best configuration
for $V_m$, $V_a$ and $P_g$. The exception is $Q_g$, whose best configuration
additionally includes $\mathcal{L}_{\mathrm{att}}$. We report this rather than
interpret it: the $Q_g$ column spans only $33.746$ to $35.240$ across the eight
distilled configurations, a range narrow enough that the ordering within it
should not carry an architectural conclusion on a single seed.

\paragraph{Reactive power falls below the teacher.}
All eight distilled configurations achieve a lower $Q_g$ error than the teacher
($33.746$--$35.240$ against $35.692$), while the undistilled baseline is far
worse at $47.236$; the effect is therefore attributable to distillation rather
than to the reduced architecture. A student surpassing its teacher on a subset
of outputs is a documented outcome of
distillation~\citep{stanton2021doeskdwork,beyer2022patientconsistent}: the
teacher's predictions are a smoother regression target than the AC-OPF labels,
and a reduced-capacity model fitted to that target can generalise better on the
noisiest channel. $Q_g$ is the plausible candidate, being the least constrained
quantity in the case study, set by converter control within the wide IEEE~1547
band rather than pinned by the power-flow equations. Isolating the mechanism
would require varying student capacity, which we have not done.

\paragraph{Fidelity to the teacher selects different objectives.}
The configuration that best matches ground truth is not in general the one that
best imitates the teacher. The two agree for $V_m$, where output plus latent
distillation minimises both, and disagree for dispatch: the lowest
teacher-referenced error for $P_g$ comes from output distillation alone and for
$Q_g$ from output plus attention distillation, neither of which is the most
accurate configuration for that quantity. Relative to the baseline, the best
configurations reduce teacher-referenced error by $57.4\%$ on $V_m$, $56.2\%$ on
$P_g$ and $80.8\%$ on $Q_g$, all larger than the corresponding gains against
ground truth. The added terms therefore buy fidelity more readily than
accuracy, and the two should be reported separately, as
\citet{stanton2021doeskdwork} argue for classification.

\paragraph{Marginal effect of each term.}
Table~\ref{tab:distillation-main-effects} isolates each term by averaging the
change in error over every pair of configurations that differ only in that
term. Output distillation dominates by one to two orders of magnitude: it
accounts for $+19.31$ of the roughly $+19.5$ total improvement available on
$P_g$ and $+12.25$ on $Q_g$. Its entry is estimated from the single
baseline-to-$\mathcal{L}_{\mathrm{out}}$ contrast, since no configuration
without output distillation was trained, which is why no spread is reported for
it; the remaining three terms each average four contrasts.

Against that, the three structural terms are second-order. Physics-residual
distillation is negative on all four quantities, and is the only term whose
mean exceeds its own spread in all four columns, making it the one clearly
resolved effect besides output distillation. Latent distillation improves
$Q_g$ ($+0.66 \pm 0.38$) and is negative on the other three, though only the
$Q_g$ entry is larger than its spread. Attention distillation is within its own
spread on $V_m$, $P_g$ and $Q_g$, and is resolved only on $V_a$, where it is
negative ($-0.76 \pm 0.61$). On this evidence, latent distillation has a small
reactive-power benefit, attention distillation has no effect we can separate
from run-to-run variation, and the physics-residual term is harmful.

\begin{table}[h]
\centering
\small
\begin{tabular}{l|cccc}
\toprule
\textbf{Loss term} & $\Delta V_m$ & $\Delta V_a$ & $\Delta P_g$ & $\Delta Q_g$ \\
\midrule
Output KD  & $\mathbf{+0.4272}$ & $\mathbf{+7.8847}$ & $\mathbf{+19.3056}$ & $\mathbf{+12.2514}$ \\
Physics KD & $-0.0999 \pm 0.0606$ & $-1.7491 \pm 1.0676$ & $-1.7155 \pm 1.2020$ & $-0.5585 \pm 0.4717$ \\
Latent     & $-0.0327 \pm 0.0583$ & $-0.6464 \pm 1.1335$ & $-1.0418 \pm 1.2703$ & $+0.6609 \pm 0.3840$ \\
Attention  & $+0.0078 \pm 0.0351$ & $-0.7584 \pm 0.6116$ & $-0.1895 \pm 0.5023$ & $+0.0487 \pm 0.3360$ \\
\bottomrule
\end{tabular}
\caption{Marginal impact of adding each distillation loss on the test set. Each
entry is the mean $\pm$ sample standard deviation of the change in RMSE over all
configuration pairs differing only in that term, scaled by $10^3$; positive
values indicate improved accuracy. The spread is over configurations, not over
seeds, and all runs use a single seed. Output KD is estimated from the single
baseline-to-$\mathcal{L}_{\mathrm{out}}$ contrast and so has no spread.}
\label{tab:distillation-main-effects}
\end{table}

\subsection{Discussion}
\label{sec:discussion}

\paragraph{Output matching carries the transfer.}
Matching the teacher's outputs recovers most of what is recoverable, and the
three terms that reach into its internal computation add little. This is
consistent with systematic comparisons of representation-level objectives on
graphs~\citep{joshi2024representationkdgnn}: a PF teacher's prediction is
already a dense, per-node, physically typed signal, unlike the single soft label
of a classification task, so the marginal information in its hidden states is
smaller. For this class of surrogate the simplest objective is therefore most of
the method, and the cheapest, needing no layer mapping, no projection matrices
and no access to teacher internals.

\paragraph{The physics-residual term is harmful in its aggregated form.}
The term we introduced as the novel component of the objective is the one that
consistently degrades accuracy, and the formulation in
Eq.~\ref{eq:physics-residual-summary} is, we believe, the reason. It reduces the
entire bus-level residual field to one scalar per layer, the mean norm over
buses, before matching. Two residual patterns with the same mean but different
spatial distributions are indistinguishable under this target, so the term
carries almost no information about \emph{where} the teacher's physical error
sits, which was the property it was meant to transfer. What it does carry is a
magnitude, and matching a magnitude is wrong in a specific way: wherever the
teacher's residual exceeds the student's, the term pushes the student to become
less physically consistent in order to agree, working against the absolute
Kirchhoff penalty already inside $\mathcal{L}_{\mathrm{GraphKit}}$. The layer
mapping compounds this, since residuals at an intermediate layer of a 12-layer
teacher and at the matched layer of a 6-layer student are at different stages of
refinement.

The term therefore seems worth reformulating rather than discarding. Matching
the residual \emph{vector} at each bus, tail-weighted so that a few weak or
heavily loaded buses cannot be averaged away, would preserve the spatial
information the present form discards; making the penalty one-sided, active
only where the student's residual exceeds the teacher's, would remove the
incentive to inherit teacher error. We state these as the natural next
experiment rather than as a result.

\paragraph{What these experiments do not establish.}
Two limitations bound every claim above. All configurations use a single seed,
so the $\pm$ values in Table~\ref{tab:distillation-main-effects} describe spread
across configurations rather than run-to-run variation, and the smaller effects,
particularly the attention entries and the ordering within the $Q_g$ column,
need replication. More importantly for a paper whose central term is physical,
we report no physics-residual metric, so it remains possible that
physics-residual distillation improves nodal power balance while costing RMSE, a
trade an operator might accept; adding mean, p95 and maximum nodal residual per
configuration would most strengthen the study.

\subsection{Edge-Deployment Considerations}
\label{sec:edge}

The parameter reduction in Table~\ref{tab:teacher-student-comparison} is
deterministic: the student has $32.5\times$ fewer parameters, with FP32 weight
storage falling from $76.6$~MiB to $2.35$~MiB, and reduces the message-passing
depth from 12 layers to 6. These figures establish a substantially smaller
architectural footprint, but they do not by themselves establish end-to-end
latency or peak runtime memory on an edge accelerator. In GNN inference at
batch size one, irregular gather/scatter operations and the number of
message-passing rounds can dominate arithmetic on weights
\citep{barillaro2025gnnedge}; activations, graph buffers and runtime workspaces
also contribute to peak memory. Device-level benchmarking under a fixed power
mode, synchronized timing and a defined warm-up protocol is therefore left to
a subsequent revision rather than inferred from parameter count alone.

\section{Conclusions}
\label{sec:conclusions}

We asked whether a grid foundation model can be reduced to a size appropriate
for edge inference without losing the behaviour that makes it useful, and which
parts of the teacher are worth transferring. Distilling a
12-layer, 20.1M-parameter GridFM into a 6-layer, 0.62M-parameter student on a
renewable variant of IEEE~118 recovers much of the capacity lost to the
reduction: against the same architecture trained without a teacher, error falls
by up to $48\%$, the best configuration closes three quarters of the
student-teacher gap on voltage magnitude, and on reactive-power dispatch every
distilled student is more accurate than the teacher it was distilled from.

The decomposition of the objective is the more useful result, and it is largely
negative. Output matching accounts for almost all of the transfer; latent-state
matching adds a small reactive-power improvement; attention matching adds
nothing separable from run-to-run variation; and physics-residual distillation,
the term we proposed as the contribution of the objective, degrades accuracy on
all four quantities. We attribute this to the aggregated form in which we posed
it, which collapses a bus-level residual field to one scalar per layer and so
transfers a magnitude rather than a spatial pattern, while giving the student an
incentive to reproduce the teacher's physical error rather than reduce it. A
bus-level, tail-weighted and one-sided reformulation would test the formulation
rather than the underlying idea, and is the obvious next step. That accuracy and
fidelity select different configurations is a further reason to report the two
separately here rather than treat teacher agreement as a proxy for quality.

Two limits bound these conclusions: a single seed, so the smaller effects need
replication, and the absence of a physics-residual metric, without which we
cannot say whether the physics term trades accuracy for physical consistency,
which for grid operation would be a different verdict on it. The $32.5\times$
parameter reduction and the halved depth are properties of the architecture,
and we have kept them separate from any latency claim, which for a GNN at batch
size one is governed by memory access rather than weight count and must be
established on the device (Section~\ref{sec:edge}). Beyond that, the natural
continuations are the reformulation above, stacking quantization and
hardware-aware pruning, which are orthogonal to
distillation~\citep{tailor2021degreequant,shen2022halp}, adaptive inference
depth~\citep{rahmath2024earlyexitsurvey}, and moving from state estimation to
control, which would require a safety mechanism and closed-loop evaluation.

\subsection*{AI use statement}
In this work, we used generative AI tools (Claude, Anthropic) to assist with
the literature search, with drafting and editing the text of the Abstract,
Introduction, State of the Art, the discussion of results and the Conclusions,
for formatting tables, for arithmetic checks on values computed by the authors
from experimental logs, and for proofreading. We have not used generative AI
tools to generate research ideas, data, experiments or results. All AI-suggested references were verified by the authors against
the original publications, and all AI-assisted text was reviewed and revised
by the authors. We take responsibility for the final content of this work,
including text, claims or artifacts produced with the aid of generative AI.

\bibliography{gridfmbibliography}
\bibliographystyle{plainnat}

\appendix
\section{Notation}
\label{app:notation}

{\small
\begin{tabular}{ll}
$f_T$ & Frozen teacher model. \\
$f_S$ & Student model trained by distillation. \\
$\mathcal{V}_b$ & Set of bus nodes in the batch graph. \\
$\mathcal{V}_g$ & Set of generator nodes in the batch graph. \\
$\mathcal{V}_{\tau}$ & Node set for node type $\tau$. \\
$\mathcal{T}$ & Set of matched node types, typically $\{\mathrm{bus}, \mathrm{gen}\}$. \\
$\mathcal{R}$ & Set of graph relation types used for attention matching. \\
$\mathcal{E}_r$ & Edge set for relation type $r$. \\
$\mathcal{M}$ & Set of matched student-teacher layer pairs $(\ell_S,\ell_T)$. \\
$\ell_S$ & Student layer index. \\
$\ell_T$ & Teacher layer index. \\
$\hat{\mathbf{y}}^{\,b}_{M,i}$ & Bus output predicted by model $M$ for bus node $i$. \\
$\hat{\mathbf{y}}^{\,g}_{M,j}$ & Generator output predicted by model $M$ for generator node $j$. \\
$\mathbf{h}^{\tau,(\ell)}_{M,i}$ & Hidden state of model $M$ at layer $\ell$ for node $i$ of type $\tau$. \\
$P_{\tau}^{(\ell_S,\ell_T)}$ & Trainable projection from student latent dimension to teacher latent dimension. \\
$r^{P,(\ell)}_{M,i}$ & Active-power residual for model $M$ at layer $\ell$ and bus $i$. \\
$r^{Q,(\ell)}_{M,i}$ & Reactive-power residual for model $M$ at layer $\ell$ and bus $i$. \\
$\rho_M^{(\ell)}$ & Mean bus-level active/reactive residual norm for model $M$ at layer $\ell$. \\
$\bar{\alpha}^{r,(\ell)}_{M,e}$ & Head-averaged Transformer attention weight for model $M$, relation $r$, edge $e$, layer $\ell$. \\
$\lambda_{\mathrm{out}}$ & Weight for output distillation. \\
$\lambda_{\mathrm{phys}}$ & Weight for physics-residual distillation. \\
$\lambda_{\mathrm{lat}}$ & Weight for latent-state distillation. \\
$\lambda_{\mathrm{att}}$ & Weight for attention distillation. \\
$\mathcal{L}_{\mathrm{GraphKit}}$ & Standard GraphKit training loss, including supervised and physics-informed terms. \\
\end{tabular}
}

\end{document}